\pdfoutput=1
\documentclass[11pt,a4paper]{article}
\usepackage[hyperref]{naaclhlt2019}
\usepackage{times}
\usepackage{latexsym}
\usepackage{url}
\usepackage{amssymb}
\usepackage{amsmath}
\usepackage{graphicx}
\usepackage{booktabs}
\usepackage{mathtools}
\usepackage{fancyhdr}
\usepackage{xcolor}
\usepackage{natbib}
\usepackage[utf8]{inputenc}
\usepackage{amsmath}
\usepackage{stmaryrd}
\usepackage{pst-node}
\usepackage{tikz-cd} 
\usepackage{multicol}
\usepackage{sidecap}
\usepackage{subcaption}

\DeclareMathOperator*{\argmax}{arg\,max}

\aclfinalcopy 
\def\aclpaperid{964} 

\title{An Incremental Iterated Response Model of Pragmatics}

\author{Reuben Cohn-Gordon \\
  Stanford \\\And
  Noah D. Goodman \\
  Stanford \\
  {\tt \{reubencg,ngoodman,cgpotts\}@stanford.edu} \\\And
  Christopher Potts \\
  Stanford \\}

\date{}

\newcommand{\Speaker}{S}
\newcommand{\Listener}{L}

\newcommand{\eg}[1]{(\ref{#1})}

\newcommand{\sem}[1]{\llbracket #1 \rrbracket}

\newcommand{\SONESENTGP}{\Speaker_{\text{1}}^{\text{UTT-GP}}}
\newcommand{\SONESENTLP}{\Speaker_{\text{1}}^{\text{UTT-IP}}}
\newcommand{\LZEROSENT}{\Listener_{\text{0}}^{\text{UTT}}}
\newcommand{\LONESENTGP}{\Listener_{\text{1}}^{\text{UTT}}}

\newcommand{\LZEROWORD}{\Listener_{0}^{\text{WORD}}}
\newcommand{\SONEWORD}{\Speaker_{1}^{\text{WORD}}}
\newcommand{\LONEWORD}{\Listener_{1}^{\text{WORD}}}

\newcommand{\cost}{\mathit{cost}}

\newcommand{\secref}[1]{section~\ref{#1}}

\newcommand{\Figref}[1]{Figure~\ref{#1}}
\newcommand{\figref}[1]{figure~\ref{#1}}

\newcommand{\word}{\emph{word}}
\newcommand{\stoptoken}{\textsc{stop}}

\begin{document}
\maketitle
\begin{abstract}
Recent \emph{Iterated Response} (IR) models of pragmatics conceptualize language use as a recursive process in which agents reason about each other to increase communicative efficiency. These models are generally defined over complete utterances. However, there is substantial evidence that pragmatic reasoning takes place incrementally during production and comprehension. We address this with an incremental IR model. We compare the incremental and global versions using computational simulations, and we assess the incremental model against existing experimental data and in the TUNA corpus for referring expression generation, showing that the model can capture phenomena out of reach of global versions.
\end{abstract}

\section{Introduction}\label{sec:intro}

A number of recent Bayesian models of pragmatics conceptualize language use as a recursive process in which abstract speaker and listener agents reason about each other to increase communicative efficiency and enrich the meanings of the utterances they hear in context-dependent ways  (\citealt{Jaeger:2007,Jaeger:2011,Franke09DISS,Frank:Goodman:2012}; for overviews, see \citealt{Franke:Jaeger:2014,goodman2016pragmatic}). For example, in these models, pragmatic listeners reason, not about the literal semantics of the utterances they hear, but rather about pragmatic speakers reasoning about simpler listeners that are defined directly in terms of the literal semantics. In this back-and-forth, many phenomena characterized by \citet{Grice75} as \emph{conversational implicatures} emerge naturally as probabilistic inferences.

In general, these \emph{iterated response} (IR) models separate pragmatic reasoning from incremental processing, in that the calculations are done in terms of complete utterances. However, there is substantial evidence that pragmatic processing is incremental: listeners venture pragmatic inferences over the time-course of the utterances they hear, which influences the choices that speakers make. To address this, we develop an IR model that is incremental in the sense that pragmatic reasoning takes place word-by-word (though the process could be defined in terms of different linguistic units, like morphemes or phrases).

A variant of this model was applied successfully to \emph{pragmatic image captioning} by \citet{cohn2018pragmatically}; here we concentrate on its qualitative behavior and linguistic predictions. We present computational experiments which demonstrate that incremental and global pragmatics make different predictions, and we show that a speaker that incrementally makes pragmatically informative choices arrives at an utterance which is globally informative. We then argue that an incremental model can account for two empirical observations out of reach of a global model: (i) the asymmetry between adjective--noun and noun--adjective languages in over-informative referential behavior \citep{rubio2016redundant}, and (ii) the anticipatory implicatures arising from contrastive modifiers \citep{sedivy2007implicature}. The first of these observations requires a model of language production, while the second requires a model of language interpretation, and as such these case studies serve to demonstrate both aspects of incremental pragmatics. Finally, we apply the model to the TUNA corpus for referring expression generation \citep{Gatt-etal:2009:REG}, showing that it makes more realistic predictions about attributive modifiers than does its global counterpart.

\section{Iterated Response Models}

We construct our model within the \emph{Rational Speech Acts} (RSA) paradigm \citep{Frank:Goodman:2012,Goodman:Stuhlmuller:2013}. RSA and its extensions have been applied to a wide range of pragmatic phenomena, including scalar implicatures \citep{Frank-etal:2016,Potts-etal:2016}, manner implicatures \citep{Bergen:Levy:Goodman:2014}, hyperbole \citep{Kao-etal:2014}, metaphor, and politeness \citep{Yoon-etal:2016}. In addition, RSA can be cast as a machine learning model, thereby allowing us to study pragmatic reasoning in large corpora and complex environments \citep{Vogel-etal:2013,Monroe:Potts:2015,Monroe:Hawkins:Goodman:Potts:2017,D16-1125}.


Standard RSA models are global in the sense that the pragmatic reasoning is defined over complete utterances. Speakers are conditional distributions of the form $P(u|w)$, while listeners are of the form $P(w|u)$, for an utterance $u$ and state $w$. We first present this global formulation (\secref{global}), and then we show how to reformulate it so that utterances are sequences of linguistic units $u = [u_{1}, \ldots, u_{n}]$ and the core RSA reasoning is applied to each step $u_{i}$ given $[u_{1}, \ldots u_{i-1}]$.

\Figref{fig2} presents a running illustrative example. We imagine there are three referents, a red dress (R1), a blue dress (R2), and a red hat (R3). We have a simple language composed of three utterances, \emph{dress}, \emph{red dress}, and \emph{red object}, each with its expected semantics. For the RSA calculation, we make the background assumption that the speaker and listener are playing a coordination game: they succeed to the extent that the listener can use the speaker's utterance to identify the speaker's intended referent.

\subsection{Global Pragmatics} \label{global}

We define our global RSA agents as follows:
\begin{align}
\LZEROSENT(w | u)  & \propto \sem{u}(w)\label{L0} \\
\SONESENTGP(u | w) & \propto e^{\log(\LZEROSENT(w | u))-\cost(u)}\label{SGP}\\
\LONESENTGP(w | u) & \propto \SONESENTGP(u | w)\label{LGP}
\end{align}

Here, $\sem{\cdot}$ is an interpretation function mapping utterances to functions from referents to $\{0,1\}$. Thus, the literal listener $\LZEROSENT$ is simply a probabilistic version of the truth conditions established by $\sem{\cdot}$; given an utterance $u$, $\LZEROSENT$ evenly distributes probability mass to the worlds compatible with $u$ according to $\sem{\cdot}$.

The pragmatic speaker $\SONESENTGP$ is more sophisticated than a literal agent, in that $\SONESENTGP$ reasons about $\LZEROSENT$, taking message costs into account. We take $\cost$ to be a language model, which could either be estimated from data (higher probability to attested utterances), derived from a grammar (higher probability to grammatical utterances), or simply assign longer utterances more cost. Intuitively, $\SONESENTGP$ prefers utterances which are not only true but best convey to $\LZEROSENT$ which world the speaker is in. This is illustrated in \figref{fig2:global}: whereas all three messages are true of R1, $\SONESENTGP$ prefers \emph{red dress} because it is the most specific. In this sense, $\SONESENTGP$ is a model of a Gricean informative speaker.

The pragmatic listener $\LONESENTGP$ in turn reasons about what world state $\SONESENTGP$ must be in such that the observed utterance was chosen, and thus draws more refined inferences than $\LZEROSENT$. We see this in \figref{fig2:global} as well, with respect to \emph{dress} and \emph{red object}. Whereas $\LZEROSENT$ regards these messages as completely ambiguous, $\LONESENTGP$ (softly) disambiguates them: \emph{dress} is heavily biased toward R2, and \emph{red object} is heavily biased toward R3.  This inference formalizes the intuitive reasoning that if the speaker of \emph{red object} had been referring to R1, they would have used the more specific, informative \emph{red dress}; their avoidance of this means they must be referring to R3.

\newcommand{\exlistener}[4]{
\setlength{\tabcolsep}{2pt}
$\begin{array}{@{} r r r r @{}}
\toprule
#1 & \text{R1} & \text{R2} & \text{R3} \\
\midrule
\emph{dress}      & #2 \\
\emph{red dress}  & #3 \\
\emph{red object} & #4 \\
\bottomrule
\end{array}$}

\newcommand{\exspeaker}[4]{
\setlength{\tabcolsep}{2pt}
$\begin{array}{@{} r r r r @{}}
\toprule
#1 & \emph{dress} & \emph{red dress}  & \emph{red object} \\
\midrule
\text{R1} & #2 \\
\text{R2} & #3 \\
\text{R3} & #4 \\
\bottomrule
\end{array}$}

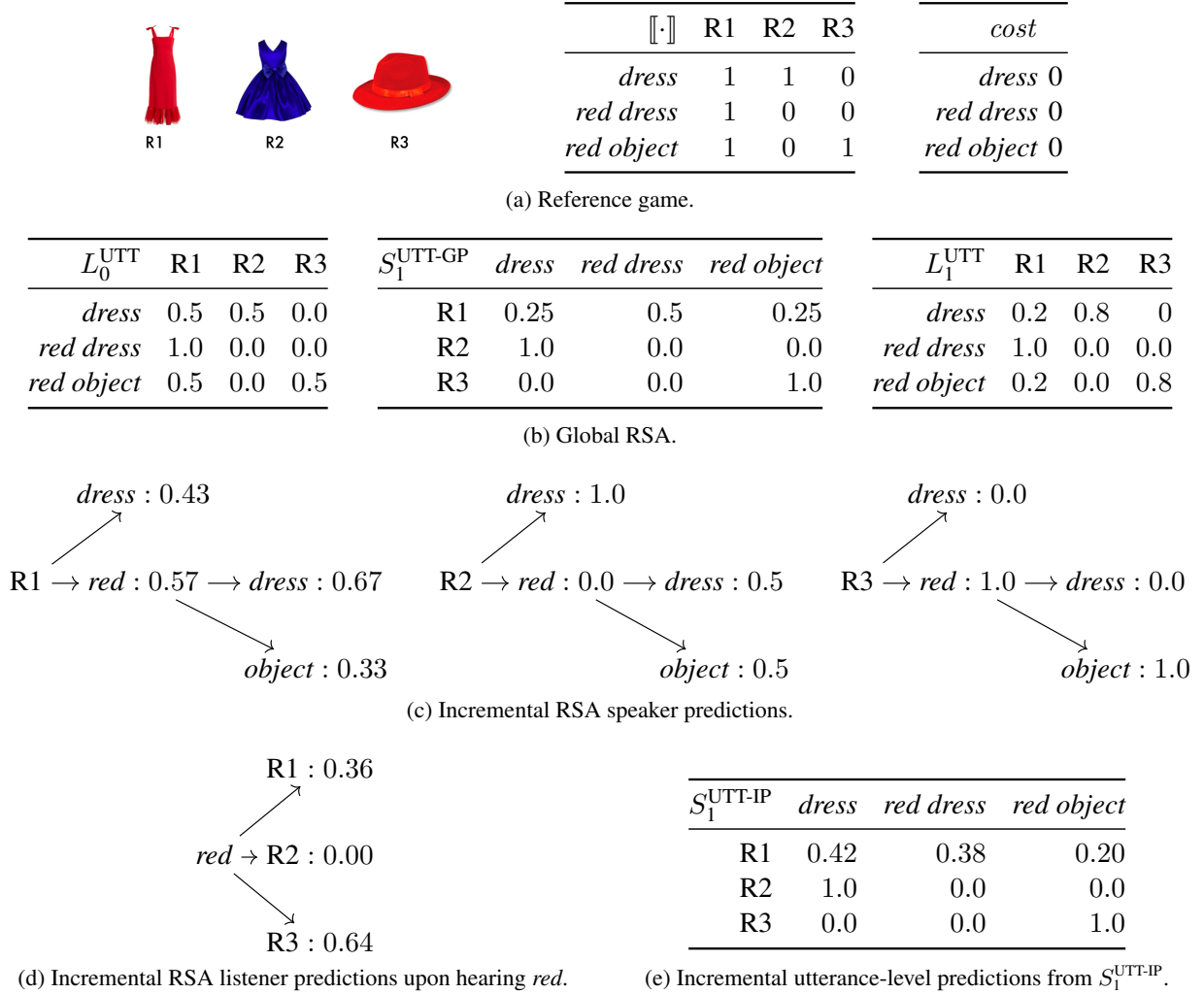
\begin{figure*}[t]
\begin{subfigure}{1\textwidth}
\centering
\parbox{0.3\textwidth}{\includegraphics[width=0.28\textwidth]{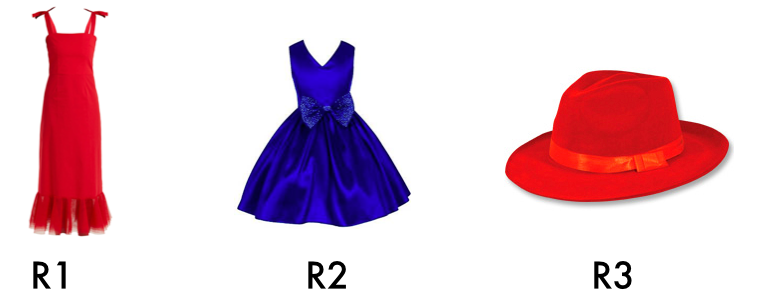}}
\qquad 
\exlistener{\sem{\cdot}}{1 & 1 & 0}{1 & 0 & 0}{1 & 0 & 1}
\qquad
\begin{tabular}{r r}
\toprule
$\cost$ & \\
\midrule
\emph{dress}      & 0 \\
\emph{red dress}  & 0 \\
\emph{red object} & 0 \\
\bottomrule
\end{tabular}
\caption{Reference game.}\label{fig2:game}
\end{subfigure}

\vspace{8pt}

\begin{subfigure}{1\textwidth}
\centering
\exlistener{\LZEROSENT}{0.5 & 0.5 & 0.0}{1.0 & 0.0 & 0.0}{0.5 & 0.0 & 0.5}
\quad
\exspeaker{\SONESENTGP}{0.25 & 0.5 & 0.25}{1.0 & 0.0 & 0.0}{0.0 & 0.0 & 1.0}
\quad
\exlistener{\LONESENTGP}{0.2 & 0.8 & 0}{1.0 & 0.0 & 0.0}{0.2 & 0.0 & 0.8}
\caption{Global RSA.}\label{fig2:global}
\end{subfigure}

\vspace{12pt}

\begin{subfigure}{1\textwidth}
\centering
\begin{tikzcd}[cramped, column sep=tiny]
& \emph{dress}: 0.43 & \\
\text{R1}\arrow[r]\arrow[ur] & \emph{red}: 0.57 \arrow[r]\arrow[rd] & \emph{dress}: 0.67 \\
                             &                                    & \emph{object}: 0.33
\end{tikzcd}
\hfill
\begin{tikzcd}[cramped, column sep=tiny]
& \emph{dress}: 1.0 & \\
\text{R2}\arrow[r]\arrow[ur] & \emph{red}: 0.0 \arrow[r]\arrow[rd] & \emph{dress}: 0.5 \\
                             &                                    & \emph{object}: 0.5
\end{tikzcd}
\hfill
\begin{tikzcd}[cramped, column sep=tiny]
& \emph{dress}: 0.0 & \\
\text{R3}\arrow[r]\arrow[ur] & \emph{red}: 1.0 \arrow[r]\arrow[rd] & \emph{dress}: 0.0 \\
                             &                                    & \emph{object}: 1.0
\end{tikzcd}
\caption{Incremental RSA speaker predictions.}\label{fig2:incremental:speaker}
\end{subfigure}

\vspace{12pt}

\begin{subfigure}[b]{0.48\textwidth}
\centering
\begin{tikzcd}[cramped, column sep=tiny]
& \text{R1}: 0.36 & \\
\emph{red}\arrow[r]\arrow[ur]\arrow[dr] & \text{R2}: 0.00 \\
& \text{R3}: 0.64
\end{tikzcd}
\caption{Incremental RSA listener predictions upon hearing \emph{red}.}\label{fig2:incremental:listener}
\end{subfigure}
\hfill
\begin{subfigure}[b]{0.48\textwidth}
\centering
\exspeaker{\SONESENTLP}{0.42 & 0.38 & 0.20}{1.0 & 0.0 & 0.0}{0.0 & 0.0 & 1.0}
\caption{Incremental utterance-level predictions from $\SONESENTLP$.}\label{fig2:incremental:utt}
\end{subfigure}
\caption{Illustrative example comparing global and incremental RSA. For ease of comparison to the global model, we do not depict the \stoptoken\ token for the incremental model.}\label{fig2}
\end{figure*}

\subsection{Incremental Pragmatics}\label{incremental}

In natural language, speakers and listeners produce and comprehend utterances segment by segment. For present purposes, we define this process at the word level, but we emphasize that the proposed approach extends both to sub-word segments \citep{cohn2018pragmatically} and to larger syntactic units.

To approximate this incremental process, we represent utterances as sequences of words and allow RSA-style reasoning to happen at the point of production or comprehension of each word, in the order they are uttered. We represent the end of an utterance as a \stoptoken\ token, so that the choice of \stoptoken\ as the next ``word'' represents the decision that the utterance is complete.

Roughly speaking, we want to define listener models $P(w|\emph{word},c)$ and speaker models $P(\emph{word}|w,c)$, where $c$ is a sequence of words constituting the utterance so far. In order to do this, we first must define an incremental semantics. This incremental semantics is defined in terms of a global semantics and the set of available complete utterances. For any partial sequence $c$ and set of referents $W$, $\sem{c}(w) \in [0, 1]$ is the number of full-utterance extensions of $c$ true in $w$ divided by the number of possible extensions of $c$ into full utterances that are true of any world in $W$. Where $c$ is a full utterance, $\sem{c}(w) \in \{0, 1\}$ is as in the global model; where $c$ is a partial utterances, $\sem{c}$ represents the biases created by $c$.

This is not the only possible way to define an incremental semantics. One alternative is to define a probabilistic $S_0$ or $L_0$  directly (as in \citealt{cohn2018pragmatically}). We choose the method defined above since it is as close as possible to a standard truth-conditional semantics and thus permits direct comparison. Furthermore, since our incremental semantics can be generated from an utterance-level semantics, only the latter needs to be stipulated.


Just as the standard RSA model presented in \secref{global} uses as global semantics to define successive speakers and listeners, we can now define an incremental literal listener $\LZEROWORD$ and incremental pragmatic speaker and listener $\SONEWORD$ and $\LONEWORD$:
\begin{gather}
\LZEROWORD(w| c, \word) \propto \sem{c+\word}(w) \\
\begin{multlined}[t]
\SONEWORD(\word |c, w) \propto \\ \shoveright{e^{\log(\LZEROWORD(w | c, \word)) - \cost(\word)}}
\end{multlined} \\
\LONEWORD(w| c, \word) \propto \SONEWORD(\word |c, w)\label{LONEWORD}
\end{gather}

\Figref{fig2:incremental:speaker} summarizes the reasoning of the incremental pragmatic speaker $\SONEWORD$, assuming $0$ cost on all words for simplicity. This agent prefers \emph{red} as a first word when conveying R1: $\SONEWORD(\emph{red}|c=[],w=\text{R1})=0.57$. However, if R3 is the intended referent, the agent \emph{must} begin its utterance with \emph{red} (since \emph{hat} is not an available word in this simple example). As shown in \figref{fig2:incremental:listener}, this fact allows the pragmatic listener to infer from hearing \emph{red} that the referent is most likely R3: $\LONEWORD(\text{R3}|c=[],\emph{red})=0.64$. 

There may be cases in which there is no possible true continuation of a sequence of words into a true utterance. For instance, no continuation of \emph{red} constitutes a truthful description of R2. In such situations, we say that probability is evenly distributed over all choices of word, so that $\SONEWORD(\mathit{dress}|c=[\mathit{red}],w=\mathit{R2}) = \SONEWORD(\mathit{object}|c=[\mathit{red}],w=\mathit{R2}) = 0.5$. 

\subsection{An Utterance-level Incremental Speaker} \label{uttlevelinc}

From the word level agent $\SONEWORD$, we can use the chain rule to obtain $\SONESENTLP$, an \emph{utterance-level speaker} whose values are the result of incremental pragmatic inferences:\footnote{We use $u[n]$ for the $n$th element of a list $u$, and $u[:n]$ for the sublist of $u$ up to but not including $u[n]$.}
\begin{align}
\SONESENTLP(u | w) = \linebreak 
\prod_{i=1}^{n} \SONEWORD(u_{i} | c=[u_{1} \ldots u_{i-1}],w)\label{SONESENTLP}
\end{align}
Whereas $\SONESENTGP$ in \eg{SGP} makes pragmatic calculations on the basis of whole utterances, $\SONESENTLP$ makes incremental pragmatic decisions about each choice of word, which together also give rise to a distribution over utterances. 

This allows for an efficient strategy, namely \emph{greedy unrolling}, to generate an utterance from a referent $r$. We choose the first word $\mathit{word}_{1}$ of the utterance to be $\argmax_{\mathit{word}} \SONEWORD(\mathit{word}|{w=r}, {c=[]})$, and this decision then becomes part of the context for choosing the second word: $\mathit{word}_{2}$ is $\argmax_{\mathit{word}} \SONEWORD(\mathit{word}|{w=r}, {c=[\mathit{word}_{1}]})$. And so on through the entire utterance.  

\subsection{Relating the Global and Incremental Models}

\begin{figure}
\[\begin{tikzcd}[arrows={line width=1pt}, column sep=large, row sep=large]
\mathit{full:sem} \arrow[r,green]{} \arrow[d]{} &
\LZEROSENT \arrow[r,green]{} &
\Speaker{\text{1}}^{\text{UTT-\textbf{X}}} \\
\mathit{inc:sem} 
\arrow[r,red,dashed]{} \arrow[u]{} &
\LZEROWORD \arrow[r,red,dashed]{}  & \SONEWORD \arrow[u,red,dashed]{}
\end{tikzcd}
\]
\caption{Two ways of constructing an utterance-level pragmatic speaker from a semantics. The solid green path is to obtain a literal listener over full utterances and then perform pragmatics, which gives rise to $\protect\SONESENTGP$  while the dashed red path is to obtain an incremental literal listener, use it to construct a word-level pragmatic speaker from $\protect\LZEROWORD$ and then use this to define an utterance-level pragmatic speaker, $\protect\SONESENTLP$.}\label{fig1}
\end{figure}
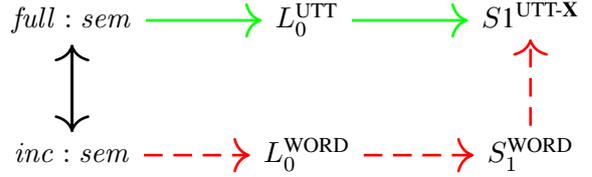

\Figref{fig1} depicts the core relationships between the global and local models, focusing on the pragmatic speaker. The agents along the solid green path define the global model of \secref{global}, while those along the dashed red path define the incremental model of \secref{incremental} as defined by $\SONESENTLP$. Importantly, while $\SONESENTGP$ and $\SONESENTLP$ are of the same \emph{type}, in the sense of being conditional probability distributions over full utterances, they are not the same distribution. 

For instance, the predictions of $\SONESENTLP$ for our illustrative example are given in \figref{fig2:incremental:utt}. Comparing them with the global pragmatic speaker predictions in \figref{fig2:global}, we see that the two make substantively different predictions. In the global model, the speaker who wishes to refer to R1 prefers \emph{red dress}. In contrast, in the incremental model, the speaker referring to R1 prefers \emph{dress}. The reason for $\SONESENTLP$ having these values is that saying \emph{dress} ensures the termination of the utterance (given the set of utterances that are available in this example), which therefore has probability $1.0$  at the next time step, while saying \emph{red} leaves two options, \emph{dress} and \emph{object}.
Thus, $\SONESENTGP$ and $\SONESENTLP$ are not only quantitatively different, but even differ in their predictions about which utterances are optimal.

\begin{figure*}
\centering
\includegraphics[width=5in,height=3in]{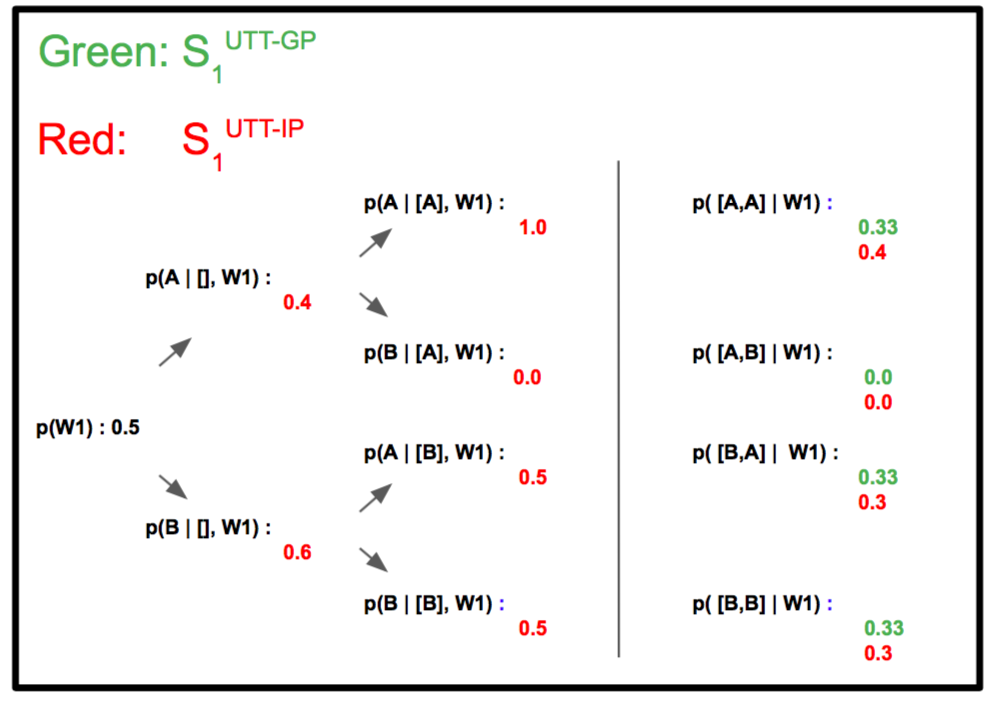} 
\caption{A depiction of the probabilities of the $\SONESENTLP$ in red and $\SONESENTGP$ in green, for a simple abstract example. The four utterances are \emph{AA}, \emph{AB}, \emph{BA} and \emph{BB}, while the two worlds are W1 and W2. The semantics assigns $u=\text{W1}$ and $w=\emph{AB}$ to 0 but all other utterance--world pairs to 1. Given a world $w$, the incremental speaker first chooses the first letter to be \emph{A} or \emph{B}, and then chooses the second letter conditioned jointly on $w$ and the first letter, to obtain a full utterance. The resulting full utterance probabilities are compared with the predictions of the global $\SONESENTGP$. As can be seen, the incremental and global speakers assign different probabilities to each utterance and are consequently distinct from each other.}
\label{example}
\end{figure*}

\Figref{example} provides an abstract example which further reinforces this difference. Moving from left to right, the numbers in green depict the $\SONEWORD$ probabilities at each of the two steps in the generation of a complete utterance, when the target reference is W1 instead of distractor W2. The probability of the full utterance at $\SONESENTLP$ is the product of the two $\SONEWORD$ steps. 

An example of the difference to $\SONESENTGP$ is shown in green. When referring to W1, $\SONESENTGP$ gives equal weight to \emph{AA}, \emph{BA} and \emph{BB}. $\SONESENTLP$, however, first chooses between \emph{A} and \emph{B}: in this decision, \emph{B} is preferred, since one of the two continuations of \emph{A}, namely \emph{AB}, is not compatible with W1. However, if \emph{A} is chosen, the subsequent choice is fully determined to be \emph{B} ($p(\emph{A}|[\emph{A}],\text{W1}=1.0$). This results in a preference for \emph{AA}. 

While we focus largely on the differences between incremental and global pragmatics, it is worth highlighting a regard in which the former behaves like the latter. 

Given a referent $r$, call an utterance $u$ \emph{weakly informative} if $\LZEROSENT(r|u) \geq \frac{1}{|W|}$, where $W$ is the set of possible referents. In other words, given $u$, the literal listener $\LZEROSENT$ will guess the correct referent with probability at least at chance (when costs are $0$). We note that the utterance $u*_r$ obtained by greedy unrolling at each step of generation, as described in \secref{uttlevelinc}, is weakly informative. To see this, observe that the $n$th word of $u*_r$ is  $\argmax_{\mathit{word}} \SONEWORD(\mathit{word}|w=r, c={u*_r[:n]})$. Since at each step $\SONEWORD$ produces a word which, at worst, does not rule out any referents for $\LZEROWORD$, the resulting sentence $u*$ at worst gives $\LZEROSENT (r|u*_r) \geq \frac{1}{|W|}$. In other words,  greedily unrolling the incremental speaker will produce an utterance which is true and has the literal listener infer the probability of the intended referent as being at least at chance.

This result suggests that the strategy of choosing the most informative word (or syntactic unit) at each point in the generation of an utterance can be used as a substitute for choosing, from all utterances, the one which is most informative. Exploring the gap in this informativity bound is an important future direction.

	One potential advantage of $\SONESENTLP$ as a plausible model of referring expression generation (over a small, discrete set of referents) is computational tractability in real-world settings where the set of possible utterances $U$ is large or unbounded. In such settings, $\SONESENTGP$ becomes intractable, owing to the normalizing term required over $U$.

	However, it is important to note that in the setup presented here, $\SONESENTLP$ is not more tractable than $\SONESENTGP$. The reason for this is that the calculation of the incremental semantics depends on factoring the global semantics: this involves an intractable normalization akin to that present in the $\SONESENTGP$. Thus, $\SONESENTLP$ is only computationally tractable if the incremental semantics in terms of which it is defined is tractable. In the present work, we define an incremental semantics in terms of a global one, in order to allow for a maximally clear comparison between the global and incremental models of pragmatics that ensue. However, an incremental semantics (or literal speaker model) can be defined or learned independently, as in \citealt{vedantam2017context} and \citealt{cohn2018pragmatically}.



\section{Application to Prior Experiments}

We now briefly consider two cases where incremental pragmatics provides an explanation of a phenomenon where global pragmatics does not seem to suffice.

\subsection{Over-informative Referring Expressions}\label{sec:spanish} 

It has been observed that, when generating referring expressions (REs), humans often provide more information than necessary to refer unambiguously \citep{engelhardt2006speakers,herrmann1976psychologie}. For instance, \citet{rubio2016redundant} shows that English speakers often use redundant color terms (e.g., \emph{the red dress}) in a scene with only a single dress, where the shorter utterance \emph{dress} would suffice. However, \citeauthor{rubio2016redundant} also notes that Spanish speakers are less likely to over-describe with the analogous referring expression, \emph{el vestido rojo}, in the same situation. This difference is a challenge for non-incremental pragmatic accounts, since, \emph{ceteris paribus}, we would expect semantically equivalent Spanish and English REs to have the same production probability.

Using incremental pragmatics, we model the English case as follows: let the referents be a red dress (R1) and a blue hat (R2), and the possible utterances be \emph{dress}, \emph{red dress}, \emph{hat}, and \emph{blue hat}, with the obvious semantics. 
 
We make the following assumption regarding the $\cost$ term: assume a cost of $1.0$ for all words but a cost of $0.0$ for the \stoptoken\ token. Further assume that an utterance's cost is the sum of the cost of its words. The effect of this cost term is to penalize longer utterances, all else being equal.
 
On these assumptions, the globally pragmatic speaker $\SONESENTGP$ prefers \emph{dress} to \emph{red dress}, since both are fully informative but the latter is costlier: $\SONESENTGP(\emph{dress}|\text{R1})=0.73 > \SONESENTGP(\emph{red dress}|\text{R1})=0.27$. Meanwhile the incremental pragmatic speaker $\SONESENTLP$ is undecided: $\SONESENTLP(\emph{dress}|\text{R1}) = \SONESENTLP(\emph{red dress}|\text{R1})=0.5$. The increase in mass on the over-informative RE \emph{red dress} in $\SONESENTLP$ as compared to $\SONESENTGP$ is the result of incremental processing: the decision between \emph{red} and \emph{dress} is made on the basis of informativity, and both words are equally informative. However, if \emph{red} is chosen, the subsequent, now over-informative word \emph{dress} has to follow, since \emph{red} on its own is not an utterance.

We explore the generality of this dynamic -- that incremental pragmatics may lead to the language model being compelled to produce longer utterances -- in section (\ref{tuna:experiment}), where we apply the model to real-world data, in the form of the TUNA corpus.

However, this effect does not obtain in Spanish, where adjectives are post-nominal. In the Spanish case, let our utterances be \emph{vestido}, \emph{vestido rojo}, \emph{sombrero}, and \emph{sombrero azul}, with the same referents and costs as before. Then there is no difference between the global and incremental models: 
\[
\begin{multlined}
\shoveleft{\SONESENTGP(\emph{vestido}|\text{R1})=0.73 >} \\ \shoveright{\SONESENTGP(\emph{vestido rojo}|\text{R1})=0.27} \\
\shoveleft{\SONESENTLP(\emph{vestido}|\text{R1})=0.73 >} \\ \shoveright{\SONESENTLP(\emph{vestido rojo}|\text{R1})=0.27}
\end{multlined}
\]
When choosing the word to follow \emph{vestido}, the incremental pragmatic speaker has no need to say \emph{rojo} rather than \stoptoken, since the goal of communicating the referent has already been completed by \emph{vestido}. As a result, the speaker chooses the less costly option, \stoptoken. The relevant difference here from the English case is that it is grammatical to stop after the first word (since the first word is a noun, not an adjective as in English).

A qualitative property which this example illustrates is a dislike in $\SONESENTLP$ for utterances which begin with a sequence of words which would mislead the incremental literal listener $\LZEROWORD$. This is the basis on which anticipatory implicatures are formed, as discussed in section \ref{anticipatory}. This is not a hard constraint: utterances which would initially mislead an incremental listener are not categorically ruled out. However, the question of whether this behavior is empirically justified is a worthwhile topic for future investigation.

\subsection{Anticipatory Implicatures} \label{anticipatory}


We now turn to a phenomenon concerning language interpretation, which we will model with an RSA listener, $\LONEWORD$.

\citet{sedivy2007implicature} provides compelling empirical evidence that humans draw pragmatic inferences partway through utterances. For instance, when shown a scene with a tall cup, a tall pitcher, a short cup, and a key, a listener who hears ``Give me the tall--'' will fixate on the tall cup before the utterance is complete. 

We take this as evidence for an incremental pragmatic listener $\LONEWORD$ which can calculate an implicature by reasoning that, had the speaker intended to refer to the pitcher, they would not have had any motivation to say ``tall''. By contrast, on the assumption that the speaker's referent is the tall cup, the contrastive modifier serves to distinguish the intended referent from the short cup. 

To model this implicature formally, we make the simplifying assumption that the possible utterances are \emph{tall cup}, \emph{short cup}, \emph{tall pitcher}, \emph{cup}, \emph{pitcher}, and \emph{key}. For consistency with the previous example, we assume the additive $\cost$ function from \secref{sec:spanish}.

On hearing \emph{tall} as the first word of an utterance, $\LONEWORD$, the incremental pragmatic listener, can draw the following inference: the intended referent is likely to have been the tall cup, since had it been the tall pitcher, there would have been no need to use the contrastive modifier \emph{tall}: $\LONEWORD(\text{the pitcher}|c=[],\emph{tall})=0.4$ while $\LONEWORD(\text{the tall glass}|c=[],\emph{tall})=0.6$.

This implicature is cancelable, and indeed, were the next word to be \emph{pitcher}, we would exclude all referents but the pitcher. In this respect, the model represents the confusion created by uttering ``tall pitcher'', where after the first word of the utterance, the majority of probability mass is on a referent (\emph{the tall cup}) which, after the second word, has no probability mass.



\section{Experimental Validation with TUNA} \label{tuna:experiment}

In order to observe the behavior of our incremental pragmatic model on real data, we make use of the TUNA corpus \cite{van2006building}. TUNA is built around a referring expression task grounded in images. The images are coded using a fixed set of attributes, and the human-produced utterances are coded using the same attributes. Thus, TUNA lets us study the core content of naturally produced referring expressions without forcing us to confront the full complexity of natural language.

Our goal is to show that, when a cost is imposed which prefers shorter utterances, the incremental model $\SONESENTLP$ is less affected, and on average produces more two-word utterances than $\SONESENTGP$. 
 
We hypothesize this on the basis of the preference of $\SONESENTLP$ for utterances where the choice of each word is made with high certainty. This means that informative one-word utterances which have reasonable probability of being extended with a second word will score lower than two-word utterances where the choice of the first word all but fixes the choice of the second. Since most one-word utterances admit the possibility of an extension to a second word, this dynamic would result in a preference for the longer, two-word utterances. This would provide further evidence that the dynamic described in section \ref{sec:spanish} generalizes from an idealized example to real language.
 
\subsection{Data}
 
The TUNA corpus defines a reference game in the sense of \figref{fig2:game}. Each \emph{trial} contains a set of images (entities), of which one or several are the target, and a human-generated referring expression for the target in the context of all the images. We refer to the full set of target and non-target entities as the \emph{context set}. 

\begin{figure}
\includegraphics[width=3in]{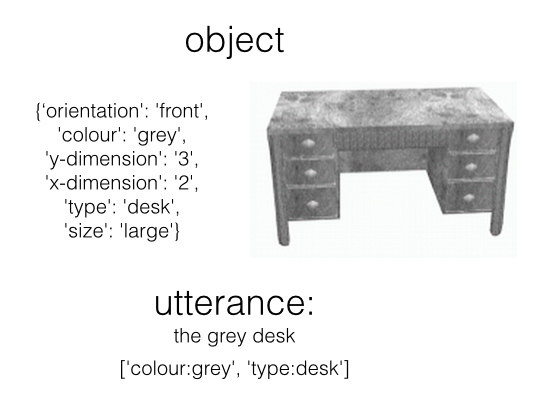} 
\caption{An example target entity from the \emph{furniture} domain, along with its coding as a dictionary, and the human generated referring expression for it in a context of other images.}
\label{fig3}
\end{figure}

Both images and utterances are coded as sets of attributes (\figref{fig3}). This coding defines a semantics. For instance, in \figref{fig3}, the utterance ``the grey desk'' is true of the entity, since \texttt{type:desk} and \texttt{colour:grey} are included in its attributes. For the \emph{furniture} domain, attributes such as color, object type, and size are coded. The \emph{people} domain is more complex, coding for more attributes, including age, clothing, hair color, glasses, and orientation. Both domains also code for the position of the image relative to the other images in the context set.


\subsection{Methods}
 
For simplicity, we restrict our model to the \emph{furniture} and \emph{people} domains where only a single referent is provided, and consider only utterances of two words or fewer. These constitute 32\% of the total utterances in the single referent corpora, and to our knowledge are not distinct in other ways than their length.
 
For each trial, the possible utterances are those from the set of all two-word utterances across the entire corpus (either of furniture or people) which are compatible with at least one of the entities in the trial. We predict the set of optimal utterances (since there may be more than one utterance with maximum probability) for both  $\SONESENTGP$ and $\SONESENTLP$. For our cost function, we assume all words have a cost of $1.0$ except  the \stoptoken\ token, which has cost $0.0$. Utterances cost the sum of their words. This has the effect of penalizing longer utterances. 

For each trial, we have a set of entities as referents, with the designated target identified among these entities. In addition, we can define the set of all possible true utterances for a given trial. Thus, it is possible to make predictions according to both $\SONESENTGP$ and $\SONESENTLP$ for each trial without having to enrich the TUNA dataset in any way.

\subsection{Results}

As expected, we find a preference for longer utterances; out of the 114 \emph{people} trials, $\SONESENTGP$ identifies 120 two-word utterances as optimal, compared to 287 for $\SONESENTLP$. In the 83 trials of the \emph{furniture} domain, $\SONESENTGP$ marks 88 two word utterances as optimal, compared to 149 for $\SONESENTLP$. (More than one utterance may be optimal for a given trial, in the event that multiple utterances have the same, maximal probability of being chosen.)

An example of a representative case is the trial where the entity in \figref{fig3} is the target, and no other distractors are grey, although others are desks. In this case, both ``grey'' and ``a grey desk'' are fully informative, in the sense of only being compatible with the target. With the cost term having the effect of penalizing longer utterances, $\SONESENTGP$ chooses ``grey'' as optimal. For $\SONESENTLP$, however, neither of these utterances are optimal, because probability is divided between stopping after ``grey'' and continuing with ``desk''. Instead, the optimal utterance, ``right middle'', describes the position of the target among the images of the context set. ``Right'' is not an available full utterance (as it is not attested in the data) and so no probability mass is lost by being divided between stopping and continuing with ``middle''.

While this result offers a possible motivation for over-informative behavior, the nature of the relation is clearly nuanced -- two-word utterances are not always more informative than one-word utterances -- and merits further work. In particular, it would be desirable to use a more direct proxy for over-informativity than preference for longer utterances.



\section{Conclusion}

In summary, we have defined a formal notion of incremental pragmatics,  with respect to both production and comprehension, and shown that it differs in meaningful ways from global pragmatic behavior, at least within the RSA paradigm. The core differences are the sensitivity to word order, which varies cross-linguistically in a way which core pragmatic reasoning does not, and the ability of an incremental model to calculate implicatures partway through an utterance. We compared these models using simulations, and we assessed them using existing psycholinguistic data and using a new experiment with the TUNA corpus.

	Exploration of RSA as a machine learning model is now underway, and this is helping to show the value of pragmatic reasoning in important NLP tasks. Such work forces us to confront the fact that the idealized RSA speaker agent must reason about all possible utterances -- the normalization constant in \eg{LGP} demands this. Prior work has sought to get around this by simplifying the space of possible utterances \citep{Monroe:Potts:2015} or by sampling a small set of utterances to approximate this normalization \citep{D16-1125,Monroe:Hawkins:Goodman:Potts:2017}. Neither solution is ideal. The incremental approach offers a scalable alternative, as long as the incremental semantics is learned, as in \citealt{cohn2018pragmatically} and \citealt{vedantam2017context}.



There is also much to be done in assessing the incremental model in the context of on-line sentence processing. We have begun to identify the key properties of the model, but the degree to which these properties accord with empirical data on production and comprehension remains a largely open question.

\section{Acknowledgments}

Many thanks to the insightful comments of the reviewers, regarding both the presentation of the material and the substantive technical comments regarding the model itself.

This material is based in part upon work supported by the
Stanford Data Science Initiative and by the NSF under Grant No. BCS-1456077.

\bibliography{naaclhlt2019}
\bibliographystyle{acl_natbib}
\end{document}